\documentclass[letterpaper]{article}
\PassOptionsToPackage{table}{xcolor}
\usepackage{aaai2027}
\nocopyright
\usepackage[hyphens]{url}
\usepackage{graphicx}
\usepackage{natbib}
\usepackage{caption}
\usepackage{subcaption}
\usepackage{amsmath,amssymb,mathtools}
\usepackage{booktabs}

\usepackage{xspace}
\usepackage{tcolorbox}

\newtcolorbox{rhwbox}{
    colback=red!30,
    colframe=white,
    boxsep=2pt,
    arc=0pt
}

\newcommand{\method}{\textsc{VGER}\xspace}
\newcommand{\pcs}{\textsc{PCS-xyt}\xspace}

\newcommand{\normop}{\operatorname{Norm}}

\newcommand{\ind}{\mathbb{I}}

\title{VGER: Voxel-Guided Global Event Ranking for Event Cloud Attribution}
\author{
Youxin Jiang,\quad
Baoheng Fu,\quad
Hongwei Ren\textsuperscript{*},\quad
Xiangqian Wu
}
\affiliations{
Harbin Institute of Technology\\
\textsuperscript{*}Corresponding author
}
\begin{document}
\maketitle 

\begin{abstract}
Event cameras produce sparse and asynchronous event streams that provide rich spatio-temporal information for efficient perception.
Recent advances in event-based models have demonstrated strong performance by directly modeling asynchronous events without dense frame reconstruction.
However, identifying the event-level evidence behind their predictions is crucial for improving model transparency and reliability.
Directly adapting point-level saliency methods from point clouds provides fine-grained attribution but overlooks event-specific spatio-temporal structures.
To address this limitation, we propose Voxel-Guided Global Event Ranking (VGER), a training-free attribution framework for point-based event cloud networks. 
VGER combines event-level gradient evidence with task-aware voxel perturbation evidence, transferring regional contribution into event-level attribution scores while preserving fine-grained resolution. Furthermore, VGER introduces a unified event ranking strategy, where high-ranked events are expected to be prediction-critical and 
low-ranked events are expected to have limited influence on predictions.
We evaluate VGER on three event-based benchmarks with PointNet, PointNet++, and EventMamba. Across nine dataset-backbone settings, VGER consistently improves both high-tail and low-tail deletion performance over point-level saliency baselines.
\end{abstract}

 \section{Introduction}                
Event cameras capture sparse and asynchronous event streams, providing high temporal resolution, low latency, and high dynamic range perception~\cite{gallego2022event}. Unlike conventional cameras that rely on dense frame representations, event-based methods directly exploit asynchronous changes to preserve fine-grained temporal dynamics while reducing redundant computation~\cite{gehrig2019end,sironi2018hats}. Recently, point-based event cloud networks have emerged as an effective paradigm for event-based recognition by representing raw events as spatio-temporal point sets and processing them with point-based architectures~\cite{qi2017pointnet,qi2017pointnet++,ren2025rethinking}. These methods have demonstrated promising performance across various event-based benchmarks~\cite{amir2017low,orchard2015converting,li2017cifar10}, benefiting from their ability to preserve the intrinsic sparsity and temporal characteristics of event streams. However, as event cloud networks become increasingly powerful, their predictions are still largely determined by complex interactions among massive asynchronous events, making it difficult to understand which events provide critical evidence for model decisions~\cite{ribeiro2016should,lundberg2017unified}.

\begin{figure}[t]
\centering
\includegraphics[width=\columnwidth]{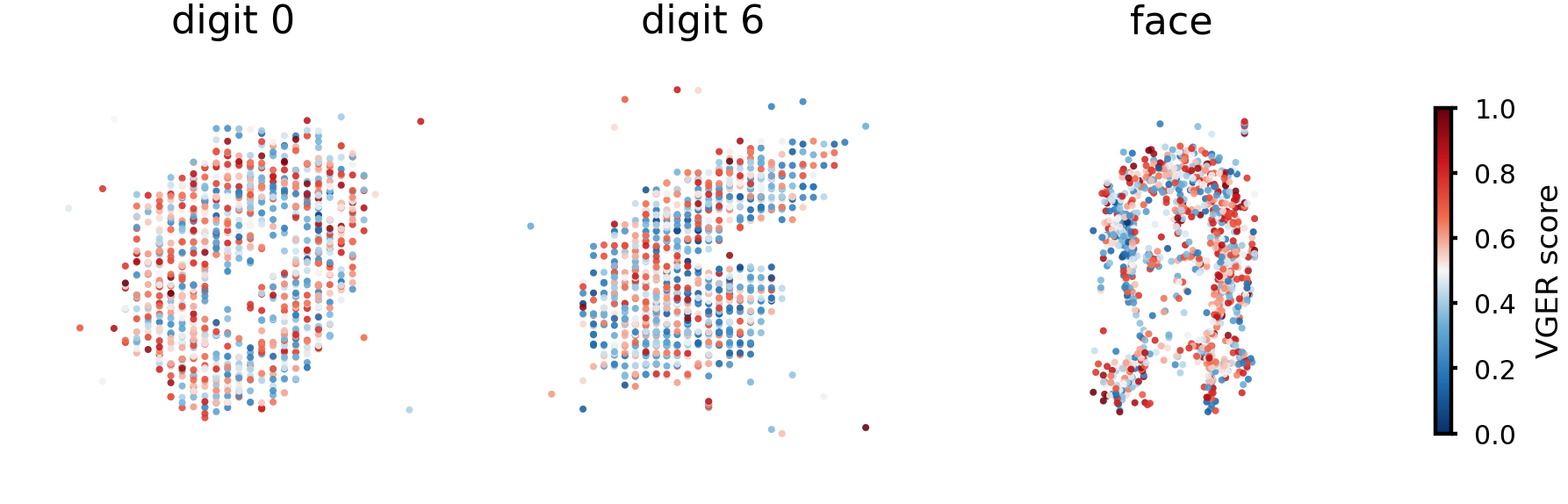}
\caption{Event-level saliency maps produced by \method on representative streams from the three evaluation datasets. Colors encode normalized saliency (red: high, blue: low). High-ranked events follow compact spatio-temporal structures that contribute to the corresponding predictions.}
\label{fig:teaser}
\end{figure}

Attribution methods for point-based models provide a natural starting point for interpreting event cloud networks~\cite{zheng2019pointcloud}. By assigning importance scores to individual points, point-level saliency approaches can identify prediction-related inputs with fine-grained spatial resolution~\cite{simonyan2013deep}. However, directly transferring these methods from conventional point clouds to event streams overlooks the unique characteristics of events. Unlike static 3D points, events are asynchronously generated, and their semantic information often emerges from local spatio-temporal structures formed by correlated event groups~\cite{gallego2022event}. For example, motion patterns and object boundaries are represented by groups of temporally correlated events rather than isolated observations. Therefore, event attribution requires not only identifying individual important events but also capturing regional structural evidence that contributes collectively to model predictions~\cite{fong2017interpretable,petsiuk2018rise}. This motivates a new attribution paradigm that bridges event-level precision and spatio-temporal structural reasoning.

A straightforward solution is to introduce regional perturbation analysis, where groups of events are removed or modified to measure their influence on model predictions~\cite{fong2017interpretable,petsiuk2018rise}. Voxel-based perturbation methods provide task-aware regional evidence by evaluating prediction changes caused by perturbing local spatio-temporal regions~\cite{deng2022voxel}. However, assigning a single importance score to all events within the same voxel inevitably sacrifices event-level discrimination. Conversely, point-level saliency preserves individual event resolution but lacks awareness of collective regional contributions. These observations reveal a fundamental granularity dilemma in event attribution: fine-grained event importance and regional structural evidence are complementary but difficult to obtain simultaneously.

To address this challenge, we propose Voxel-Guided Global Event Ranking (VGER), which combines voxel-level task evidence with event-level attribution signals to generate a more faithful global ranking of asynchronous events. VGER is a training-free attribution framework designed for point-based event cloud networks. It first derives task-aware regional evidence through voxel perturbation and transfers this evidence back to individual events. Meanwhile, event-level gradient attribution provides fine-grained sensitivity information~\cite{simonyan2013deep,zheng2019pointcloud}. By integrating these two complementary signals, VGER produces a unified event ranking that preserves both regional structural awareness and individual event resolution. Beyond conventional attribution evaluation that only examines whether important events affect predictions, we introduce a two-tail ranking perspective: highly ranked events should be prediction-critical, while low-ranked events should have limited influence~\cite{petsiuk2018rise,hooker2019benchmark}. We evaluate VGER on three event-based benchmarks with multiple point-based architectures, including PointNet, PointNet++, and EventMamba~\cite{qi2017pointnet,qi2017pointnet++,ren2025rethinking,ren2024spikepoint}. Extensive experiments demonstrate that VGER consistently improves attribution faithfulness over point-level saliency baselines across diverse dataset-backbone settings. The main contributions of this work are summarized as follows:
\begin{itemize}
    \item We introduce VGER, the first attribution framework for point-based event cloud networks, enabling fine-grained interpretation of asynchronous event streams.

    \item We propose a voxel-guided evidence fusion strategy that combines task-aware regional perturbation evidence with event-level gradient attribution to preserve both structural context and event-level resolution.

    \item We develop a global event ranking evaluation strategy with complementary high-tail and low-tail assessments, providing a more comprehensive evaluation of event attribution faithfulness.
\end{itemize}

\begin{figure*}[t]
\centering
\includegraphics[width=\textwidth]{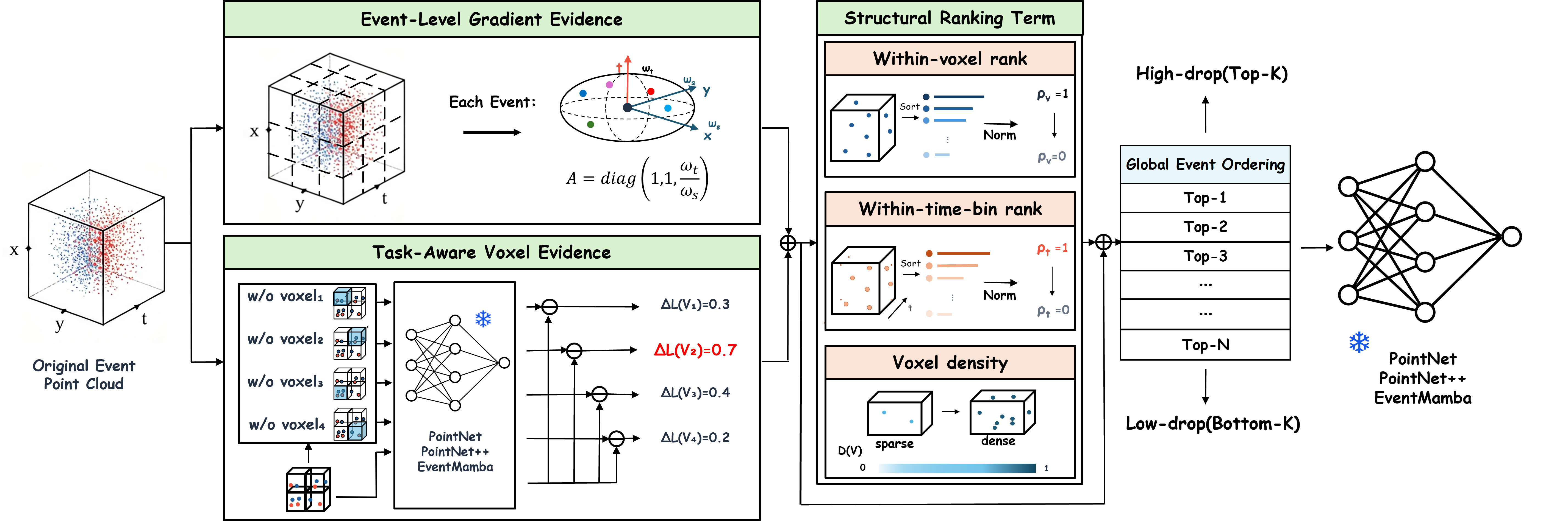}
\caption{Overview of \method. Event-level gradient evidence and task-aware voxel perturbation evidence are transferred, normalized, and fused into one global event ordering. The same ordering is evaluated at the high and low tails to distinguish prediction-critical events from events with limited influence.}
\label{fig:overview}
\end{figure*}

\section{Related Work}

\paragraph{Event representations and recognition.}
Event representations range from dense time surfaces, learned grids, and recurrent surfaces to sparse recursive, graph-based, and voxel-based forms~\cite{sironi2018hats,gehrig2019end,cannici2020differentiable,sekikawa2019eventnet,schaefer2022aegnn,deng2022voxel,ren2026systematic}. EvT exploits sparsity by retaining activated event-derived patches, while GET groups events by timestamp and polarity~\cite{sabater2022event,peng2023get}. E2PNet, PEPNet, EventMamba, and SECNet develop task-specific spatial and temporal processing for event data~\cite{lin2023e2pnet,ren2024simple,ren2025rethinking,ren2026scalable}. General point-based architectures such as PointNet, PointNet++, DGCNN, and Point Transformer provide operations that can be applied to spatio-temporal event points~\cite{qi2017pointnet,qi2017pointnet++,wang2019dynamic,zhao2021point}. These developments expand how event structure is processed across event-based tasks; \method complements them by identifying the event-level evidence behind a trained model's prediction.

\paragraph{Fine-grained and regional attribution.}
Gradient Saliency, Integrated Gradients, LRP, Grad-CAM, and SmoothGrad derive attribution from input derivatives or internal activations~\cite{simonyan2013deep,sundararajan2017axiomatic,bach2015pixel,selvaraju2017grad,chattopadhay2018grad,smilkov2017smoothgrad}. LIME fits a local interpretable surrogate, while Kernel SHAP uses a weighted linear explanation model~\cite{ribeiro2016should,lundberg2017unified}. Perturbation methods measure prediction changes under controlled input interventions~\cite{zeiler2014visualizing,fong2017interpretable,petsiuk2018rise}. For point clouds, PCS uses a differentiable center-shifting surrogate to score individual points~\cite{zheng2019pointcloud}; later work explores features, local surrogates, semantic groups, interpretable architectures, and factorized attribution maps~\cite{levi2024fast,ahmadi2024explainability,mulawade2025xai,feng2024interpretable3d,liu2024ffam}. EventPoint detects event keypoints for matching and registration~\cite{huang2023eventpoint}, and gradient attribution has also been studied for spiking-network inputs~\cite{bitar2023gradient}. The attribution methods in this group span fine-grained input evidence, group-level perturbation, feature-based explanations, and inherently interpretable architectures. In contrast, \method transfers regional task evidence to individual events before globally ranking them, directly connecting the two granularities identified in the Introduction.

\paragraph{Evaluation of explanations.}
Deletion and insertion curves evaluate an attribution ordering by tracking model behavior as inputs are removed or restored~\cite{petsiuk2018rise}. Their interpretation depends on a controlled perturbation protocol~\cite{hooker2019benchmark,tomsett2020sanity,gomez2022metrics}, while sanity checks, input-invariance and robustness tests, and infidelity/sensitivity metrics assess complementary properties of explanations~\cite{adebayo2018sanity,kindermans2019reliability,alvarez2018robustness,yeh2019fidelity}. We hold deletion cardinality and backbone-specific operators fixed across attribution methods and evaluate both ends of the same global ordering. This two-tail design tests whether high-ranked events are prediction-critical and low-ranked events have limited influence under one consistent protocol.

\section{Problem Formulation}
We consider the attribution problem for point-based event cloud networks.
An event stream is represented as a set of asynchronous events:
\begin{equation}
E=\{e_i\}_{i=1}^{N}, \qquad e_i=(x_i,y_i,t_i,p_i),
\end{equation}
where $(x_i,y_i,t_i)$ denotes the spatio-temporal coordinate and $p_i$ denotes the event polarity.
A frozen event cloud model $f_\theta$ takes $E$ as input and produces prediction logits, with $\mathcal{L}(f_\theta(E),y)$ denoting the task loss.
Given an event stream $E$ and its label $y$, the objective is to assign an importance score $S(e_i)$ to each event while keeping $\theta$ unchanged.
Unlike conventional point attribution, event attribution should capture not only individual event importance but also the contribution of local spatio-temporal structures.


For a deletion ratio $r$ and $K=\lfloor rN\rfloor$, the top- and bottom-ranked event subsets are defined as:
\begin{equation}
\begin{aligned}
D_{\mathrm{high}}(r)
&=\{e_i \mid S(e_i)\in \operatorname{TopK}(S,K)\},\\
D_{\mathrm{low}}(r)
&=\{e_i \mid S(e_i)\in \operatorname{BottomK}(S,K)\},
\end{aligned}
\label{eq:two_tail}
\end{equation}
where TopK and BottomK denote the sets of events with the largest and smallest attribution scores, respectively.
Given a unified event ranking, Eq.~\eqref{eq:two_tail} provides two complementary evaluation perspectives. Removing $D_{\mathrm{high}}$ is expected to cause larger prediction degradation, while removing $D_{\mathrm{low}}$ should have limited impact on recognition. The deletion operator $\mathcal{D}$ removes the selected events with a fixed cardinality and follows the input representation requirements of each backbone. For fair comparison, the same deletion procedure is applied to all attribution methods under each backbone.

\section{Voxel-Guided Global Event Ranking}

VGER addresses the attribution challenge through four stages.
It first derives fine-grained sensitivity from event-level gradients and then measures the task contribution of local spatio-temporal voxels. The voxel evidence is transferred back to individual events and combined with a structural term that organizes their relative ranks. Finally, the fused event scores define one global ordering for both high-tail and low-tail evaluation.

\subsection{Event-Level Gradient Evidence}

Event-level gradients provide the fine-grained component of \method. Because spatial coordinates and timestamps have different units and ranges, we normalize each $z_i=(x_i,y_i,t_i)^\top$ using a robust stream center $c=(c_x,c_y,c_t)^\top$ and a positive coordinate-wise scale $s=(s_x,s_y,s_t)^\top$:
\begin{equation}
\bar z_i=
\left(
\frac{x_i-c_x}{s_x},
\frac{y_i-c_y}{s_y},
\frac{t_i-c_t}{s_t}
\right)^\top.
\label{eq:normalized_coordinate}
\end{equation}
This normalization prevents the temporal coordinate from dominating the score solely through its numerical range.

We adapt PCS to the spatio-temporal event domain by measuring task-loss sensitivity along a differentiable center-contraction path. To distinguish temporal displacement from spatial displacement, we introduce the anisotropic metric
\begin{equation}
A_{\rho}=\operatorname{diag}(1,1,\rho),
\qquad \rho=\frac{\omega_t}{\omega_s},
\label{eq:anisotropic_metric}
\end{equation}
where $\rho$ controls the relative weight of temporal displacement. The corresponding anisotropic radius is
\begin{equation}
r_i^{\rho}=
\sqrt{\bar z_i^\top A_{\rho}\bar z_i}.
\label{eq:anisotropic_radius}
\end{equation}
We define the anisotropic center-contraction path as
\begin{equation}
\bar z_i(\tau)=
\exp(-\tau A_{\rho})\bar z_i,
\qquad \tau\geq 0.
\label{eq:anisotropic_path}
\end{equation}
Its instantaneous displacement direction is
\begin{equation}
\left.
\frac{\partial \bar z_i(\tau)}{\partial \tau}
\right|_{\tau=0}
=
-A_{\rho}\bar z_i.
\label{eq:anisotropic_direction}
\end{equation}
Therefore, the anisotropic event-level PCS score is defined as
\begin{equation}
S_{\mathrm{pcs}}^{\rho}(e_i)=
-\left\langle
\nabla_{\bar z_i}\mathcal{L},
\frac{A_{\rho}\bar z_i}{r_i^{\rho}+\epsilon}
\right\rangle
(r_i^{\rho})^{1+\alpha},
\label{eq:anisotropic_pcs}
\end{equation}
where $\mathcal{L}$ is the task loss, $\epsilon$ is a small constant, and
$\alpha$ controls the radial scaling. Eq.~\eqref{eq:anisotropic_pcs}
reduces to the original isotropic PCS formulation when $\rho=1$.

The score in Eq.~\eqref{eq:anisotropic_pcs} measures the first-order increase in task loss when an event is suppressed along the anisotropic spatio-temporal contraction direction. Adjusting $\rho$ balances temporal and spatial displacement while preserving a separate score for every event.

\subsection{Task-Aware Voxel Evidence}

Event-level gradients distinguish individual events but do not explicitly measure the finite task effect of a complete local spatio-temporal region. To obtain this complementary regional evidence, we partition the normalized event domain into $G_s\times G_s\times G_t$ voxels. Equivalently, voxelization can be performed in the anisotropically transformed coordinates
\begin{equation}
u_i=A_{\rho}^{1/2}\bar z_i,
\label{eq:anisotropic_voxel_coordinate}
\end{equation}
so that the spatial and temporal extent of each voxel follows the metric in Eq.~\eqref{eq:anisotropic_metric}.

Let $V_j$ denote an occupied voxel. We construct
$\widetilde E^{(j)}$ by moving the events in $V_j$ toward the robust stream
center while preserving the number, polarity, and temporal order of events.
The task-aware evidence of $V_j$ is
\begin{equation}
\Delta\mathcal{L}(V_j)=
\mathcal{L}(f_{\theta}(\widetilde E^{(j)}),y)
-
\mathcal{L}(f_{\theta}(E),y),
\label{eq:voxel_delta}
\end{equation}
where $f_{\theta}$ is the frozen task network and $y$ is the ground-truth
label. This voxel-level loss change is broadcast to each constituent event:
\begin{equation}
S_{\mathrm{vox}}(e_i)=
\Delta\mathcal{L}(V(e_i)).
\label{eq:voxel_score}
\end{equation}

Eq.~\eqref{eq:voxel_score} performs the transfer described in the Introduction. Rather than treating a voxel as the final attribution unit, \method assigns its measured task contribution to each constituent event. The subsequent fusion therefore retains event-level resolution while incorporating evidence about the local structure surrounding each event.

  \subsection{Structural Ranking Term}

  The event-level gradient and voxel perturbation terms address the two sides
  of the granularity dilemma. The former distinguishes individual events
  within a local region, whereas the latter captures the finite task effect of
  the region as a whole. We first combine these complementary signals into a
  base event score:
  \begin{equation}
  S_{\mathrm{base}}(e_i)=
  \beta_p\normop\!\left(S_{\mathrm{pcs}}^{\rho}(e_i)\right)
  +
  \beta_v\normop\!\left(S_{\mathrm{vox}}(e_i)\right),
  \label{eq:base_score}
  \end{equation}
  where $\normop(\cdot)$ denotes stream-wise score normalization, and
  $\beta_p$ and $\beta_v$ control the contributions of event-level gradient
  evidence and voxel-level task evidence, respectively.

  We then organize the base scores according to their local and temporal
  contexts. Let
  \begin{equation}
  \begin{aligned}
  \mathcal{E}_{V}(e_i)
  &=
  \left\{e_k\in E: V(e_k)=V(e_i)\right\},\\
  \mathcal{E}_{T}(e_i)
  &=
  \left\{e_k\in E: T(e_k)=T(e_i)\right\},
  \end{aligned}
  \label{eq:structural_event_sets}
  \end{equation}
  where $V(e_i)$ denotes the voxel containing $e_i$, and $T(e_i)$ denotes its
  temporal-bin index. Thus, $\mathcal{E}_{V}(e_i)$ contains the events in the
  same spatio-temporal voxel as $e_i$, while $\mathcal{E}_{T}(e_i)$ contains
  the events assigned to the same temporal bin.

  For an event subset $\mathcal{Q}$ containing $e_i$, let
  $r_{\mathcal{Q}}(e_i)$ be the rank of $S_{\mathrm{base}}(e_i)$ among the
  events in $\mathcal{Q}$, with larger base scores receiving larger ranks. We
  define the normalized within-set rank as
  \begin{equation}
  \rho_{\mathcal{Q}}(e_i)=
  \begin{cases}
  \dfrac{r_{\mathcal{Q}}(e_i)-1}
        {|\mathcal{Q}|-1},
  & |\mathcal{Q}|>1,\\[6pt]
  \dfrac{1}{2},
  & |\mathcal{Q}|=1.
  \end{cases}
  \label{eq:normalized_local_rank}
  \end{equation}
  The singleton case is assigned the neutral value $1/2$. The voxel-wise and
  temporal ranks are consequently given by
  \begin{equation}
  \rho_V(e_i)=
  \rho_{\mathcal{E}_{V}(e_i)}(e_i),
  \qquad
  \rho_T(e_i)=
  \rho_{\mathcal{E}_{T}(e_i)}(e_i).
  \label{eq:voxel_temporal_rank}
  \end{equation}
  Both quantities lie in $[0,1]$ and preserve the relative position of an
  event within its corresponding local context.

  We additionally characterize the occupancy of each voxel. Let
  $\mathcal{V}_{\mathrm{occ}}$ denote the set of occupied voxels and let
  $n(V_j)$ be the number of events assigned to voxel $V_j$. We define
  \begin{equation}
  \begin{aligned}
  n(V_j)
  &=
  \sum_{k=1}^{N}
  \ind\!\left[V(e_k)=V_j\right],\\
  D(V_j)
  &=
  \frac{n(V_j)}
  {\displaystyle
   \max_{V_\ell\in\mathcal{V}_{\mathrm{occ}}}n(V_\ell)}.
  \end{aligned}
  \label{eq:voxel_occupancy}
  \end{equation}
  Accordingly, $D(V_j)\in(0,1]$ for every occupied voxel. A larger value
  indicates a region containing more events, whereas $1-D(V_j)$ represents
  its relative sparsity within the current event stream.

  Using these contextual quantities, we define the structural ranking score as
  \begin{equation}
  \begin{split}
  S_{\mathrm{struct}}(e_i)&={}
  w_b S_{\mathrm{base}}(e_i)
  +
  w_c \rho_V(e_i)
  +
  w_t \rho_T(e_i)\\
  &+
  w_r\bigl(1-D(V(e_i))\bigr)
  -
  w_dD(V(e_i)).
  \end{split}
  \label{eq:struct_score}
  \end{equation}
  Here, $w_b$ retains the underlying attribution evidence, $w_c$ and $w_t$
  control the voxel-wise and temporal rank contributions, and $w_r$ and $w_d$
  calibrate the relative sparsity and occupancy of the event's voxel. The two
  occupancy components are written separately to expose their structural
  interpretations, while their combined ranking effect is governed by
  $w_r+w_d$.

  The structural term therefore refines the global event ordering without
  replacing the gradient and voxel attribution signals. The within-voxel rank
  preserves fine-grained comparisons among events receiving the same regional
  evidence, the temporal rank incorporates their relative temporal context,
  and the occupancy terms account for how events are distributed across
  spatio-temporal regions.

\subsection{Voxel-Guided Global Event Ranking}

The final \method score fuses regional, event-level, and structural evidence into one scalar for every event:
\begin{align}
S_{\mathrm{VGER}}(e_i)={}&
\lambda_v\normop(S_{\mathrm{vox}}(e_i))
+
\lambda_p\normop(S_{\mathrm{pcs}}^{\rho}(e_i))\nonumber\\
&+
\lambda_r\normop(S_{\mathrm{struct}}(e_i)).
\label{eq:vger_fusion}
\end{align}
Here, $\normop(\cdot)$ denotes stream-wise score normalization, and
$\lambda_v$, $\lambda_p$, and $\lambda_r$ control the contributions of
voxel-level, gradient-level, and structural evidence, respectively.
\method then performs a stable global ranking over the complete stream:
\begin{equation}
\pi=
\operatorname{argsort}_{e_i\in E}
S_{\mathrm{VGER}}(e_i).
\label{eq:global_ranking}
\end{equation}

\section{Experiments}

\subsection{Datasets, Backbones, and Baselines}

We evaluate DVS Gesture~\cite{amir2017low}, N-MNIST, and N-Caltech101~\cite{orchard2015converting}, the standard event-classification benchmarks alongside CIFAR10-DVS~\cite{li2017cifar10}. The backbones are PointNet~\cite{qi2017pointnet}, PointNet++~\cite{qi2017pointnet++}, and EventMamba~\cite{ren2025rethinking}. The verified baselines shared by the archived $3\times3$ matrix are Random and \pcs. A separate PointNet study compares Basic Voxel with \pcs on all three datasets.

Random uses the archived event permutations and provides a task-agnostic deletion reference. \pcs applies Eq.~\eqref{eq:anisotropic_pcs} when $\rho=1$ without voxel evidence. Basic Voxel assigns the same task-loss change to all events in one voxel; its ties are resolved at event level while respecting the exact deletion budget. We restrict the quantitative comparison to these baselines because they are the methods represented by verified, common-format curves in the supplied archive.

The PointNet, PointNet++ and EventMamba results use the full archived test counts: 6,623 for DVS Gesture, 10,000 for N-MNIST, and 17,869 for N-Caltech101. 

\subsection{Two-Tail Protocol}

All settings are compared over
\begin{equation*}
\mathcal{R}=\{0,0.05,\ldots,1.00\}.
\end{equation*}
At each ratio, exactly $\lfloor rN\rfloor$ events are deleted. Accuracy curves are summarized by trapezoidal AUC; lower is better for the high tail and higher is better for the low tail:
\begin{equation*}
\operatorname{AUC}(A)=
\sum_{k=1}^{20}\frac{A(r_k)+A(r_{k+1})}{2}(r_{k+1}-r_k).
\end{equation*}

We report exact ratio-wise wins for every setting. The two summaries answer different questions: AUC weights the magnitude of the curve difference at each ratio, whereas a win count gives every sampled ratio one vote, so a method cannot dominate either summary through a single strong region of the protocol.

\begin{table}[t]
\centering
\small
\setlength{\tabcolsep}{3.7pt}
\begin{tabular}{lcc}
\toprule
Method & High AUC $\downarrow$ & Low AUC $\uparrow$\\
\midrule
Random & 0.4977 & 0.4977\\
\pcs  & 0.4368 & 0.4934\\
\rowcolor{cyan!10}
\method & \textbf{0.3836} & \textbf{0.5404}\\
\bottomrule
\end{tabular}
\caption{\method improves both macro AUC tails over \pcs. Values are unweighted averages over nine archived settings.}
\label{tab:macro}
\end{table}

\begin{table}[t]
\centering
\scriptsize
\setlength{\tabcolsep}{3.0pt}
\begin{tabular}{llr>{\columncolor{cyan!10}}c c >{\columncolor{cyan!10}}c c}
\toprule
Backbone & Dataset & $n$ & VGER-H$\downarrow$ & PCS-H$\downarrow$ & VGER-L$\uparrow$ & PCS-L$\uparrow$\\
\midrule
PointNet & DVS Gesture & 6,623 & \textbf{0.5532} & 0.6351 & \textbf{0.8121} & 0.7781\\
PointNet & N-MNIST & 10,000 & \textbf{0.5959} & 0.6422 & \textbf{0.7339} & 0.6971\\
PointNet & N-Caltech101 & 17,869 & \textbf{0.0576} & 0.0756 & \textbf{0.2665} & 0.2180\\
PointNet++ & DVS Gesture & 6,623 & \textbf{0.4665} & 0.5311 & \textbf{0.6035} & 0.5315\\
PointNet++ & N-MNIST & 10,000 & \textbf{0.2655} & 0.2740 & \textbf{0.3811} & 0.3489\\
PointNet++ & N-Caltech101 & 17,869 & \textbf{0.0799} & 0.0817 & \textbf{0.0981} & 0.0918\\
EventMamba & DVS Gesture & 6,623 & \textbf{0.4429} & 0.5765 & \textbf{0.7369} & 0.6290\\
EventMamba & N-MNIST & 10,000 & \textbf{0.7729} & 0.8616 & \textbf{0.8935} & 0.8491\\
EventMamba & N-Caltech101 & 17,869 & \textbf{0.2180} & 0.2533 & \textbf{0.3381} & 0.2972\\
\bottomrule
\end{tabular}
\caption{\method wins both AUC tails in all nine settings. H is high-tail deletion and L is low-tail deletion; bold indicates the better AUC within each pair.}
\label{tab:full}
\end{table}


\subsection{Main Two-Tail Fidelity}

\begin{figure*}[t]
\centering
\includegraphics[width=\textwidth]{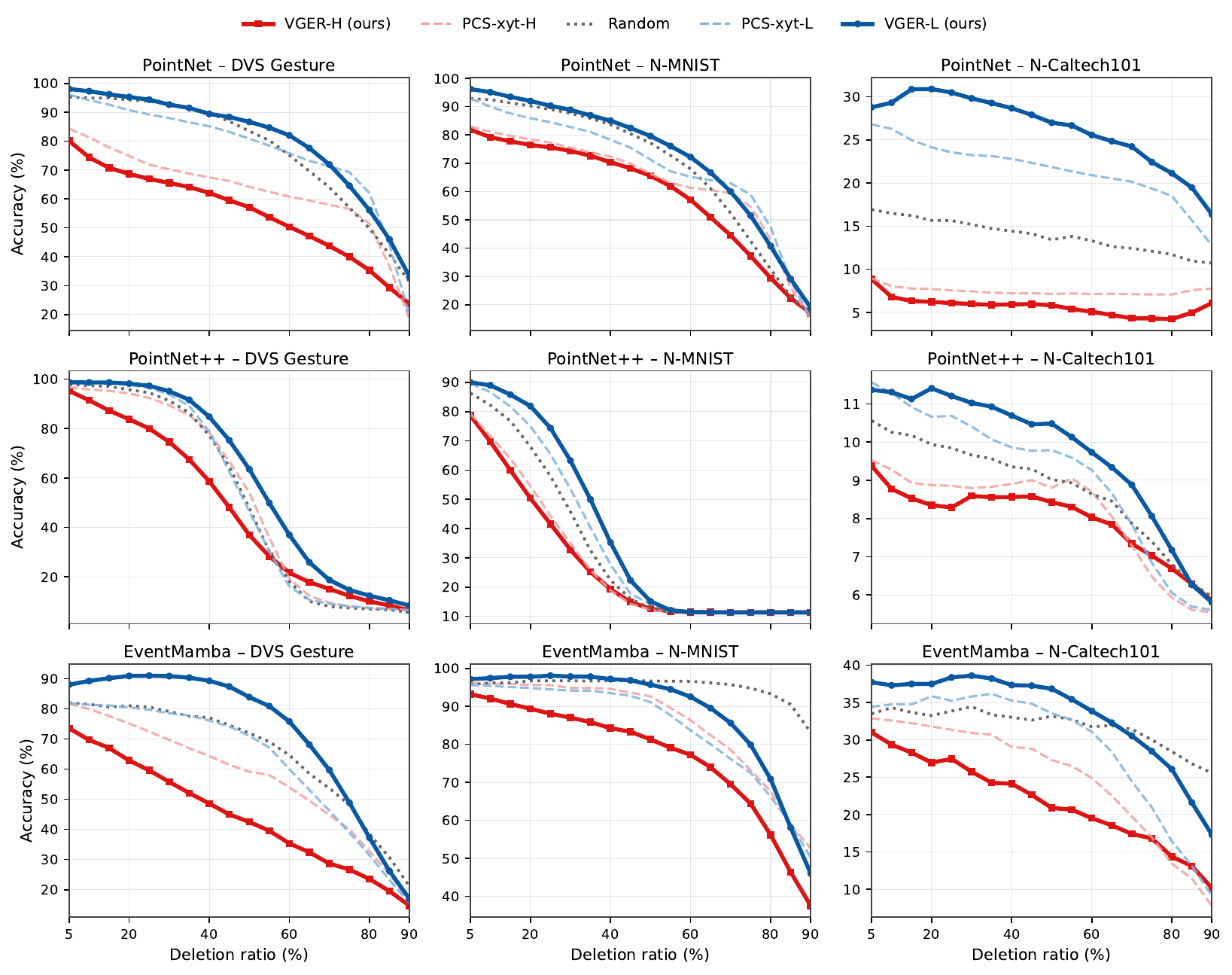}
\caption{Two-tail deletion curves for the $3\times3$ matrix of backbones and datasets. Each panel deletes $5\%$--$90\%$ of events in score order and reports accuracy. \method's high tail (VGER-H) drops below \pcs-H and Random in most settings and ratios, and its low tail (VGER-L) stays above \pcs-L in most settings and ratios, so the two \method tails separate decision-critical from decision-irrelevant events across the whole protocol.}
\label{fig:curves_3x3}
\end{figure*}

Table~\ref{tab:macro} reports the macro comparison across the archived matrix.The macro comparison shows that \method reduces high-tail AUC by 0.0532 and increases low-tail AUC by 0.0470 relative to \pcs. Its macro two-tail gap, low AUC minus high AUC, is 0.1568 compared with 0.0566 for \pcs. These averages weight each dataset--backbone setting equally. The rows behind them are uniformly directional and are all produced by frozen archived configurations, so the macro gains are not carried by any single backbone or dataset.Table~\ref{tab:full} gives the corresponding setting-level AUCs.


Figure~\ref{fig:curves_3x3} presents the main experiment of this paper: the complete two-tail deletion curves behind these counts. In most settings and deletion ratios, removing \method's top-ranked events damages accuracy earlier and deeper than removing \pcs's or Random's, whereas removing its bottom-ranked events preserves accuracy longer than \pcs's. The appendix further averages the two-tail curves over all nine settings and condenses each setting into its two-tail separation, which \method widens over \pcs in all nine settings, showing that one global ordering simultaneously isolates decision-critical and decision-irrelevant events.


Additional implementation details and supporting evidence are provided in the appendix, including the setting-specific fusion weights used in part of the experiments, the full numerical data underlying Figure~\ref{fig:curves_3x3}, and the auxiliary results from hyperparameter search and related diagnostic studies. These materials are included to make the reported two-tail fidelity results traceable beyond the summarized AUC values and to facilitate independent verification of the experimental protocol.

\subsection{Results on PointNet}
PointNet uses single max-pooling, making predictions depend on a compact critical event subset. High-tail deletion therefore tests whether \method's top-ranked events are truly decisive. \method lowers high-tail AUC from 0.6351 to 0.5532 on DVS Gesture and from 0.0756 to 0.0576 on N-Caltech101. Low-tail AUC also improves over \pcs on all datasets, with gains of 0.0340, 0.0368, and 0.0485. This consistent two-tail behavior shows reliable importance estimates for globally pooled representations.

\subsection{Results on PointNet++}
PointNet++ learns hierarchical features from nested spatial neighborhoods, so attribution depends on both local geometry and global structure. Although \pcs already has low high-tail AUCs, \method improves both tails on every dataset. On DVS Gesture, high-tail AUC drops from 0.5311 to 0.4665, while low-tail AUC rises from 0.5315 to 0.6035. Figure~\ref{fig:curves_3x3} further shows that top-ranked deletion disrupts recognition faster, whereas bottom-ranked deletion preserves accuracy longer. The smaller gains on N-MNIST and N-Caltech101 remain consistent.

\subsection{Results on EventMamba}
EventMamba models long-range temporal dependencies through selective state updates, making event-level attribution less direct than in point-wise networks. Still, \method improves both tails in all three archived runs. The largest gain appears on DVS Gesture, where high-tail AUC decreases from 0.5765 to 0.4429 and low-tail AUC increases from 0.6290 to 0.7369. N-MNIST and N-Caltech101 show the same pattern: removing highly ranked events causes sharper degradation, while removing low-ranked events better preserves accuracy.

\subsection{Ablation Study}

\begin{figure}[t]
\centering
\includegraphics[width=0.82\columnwidth]{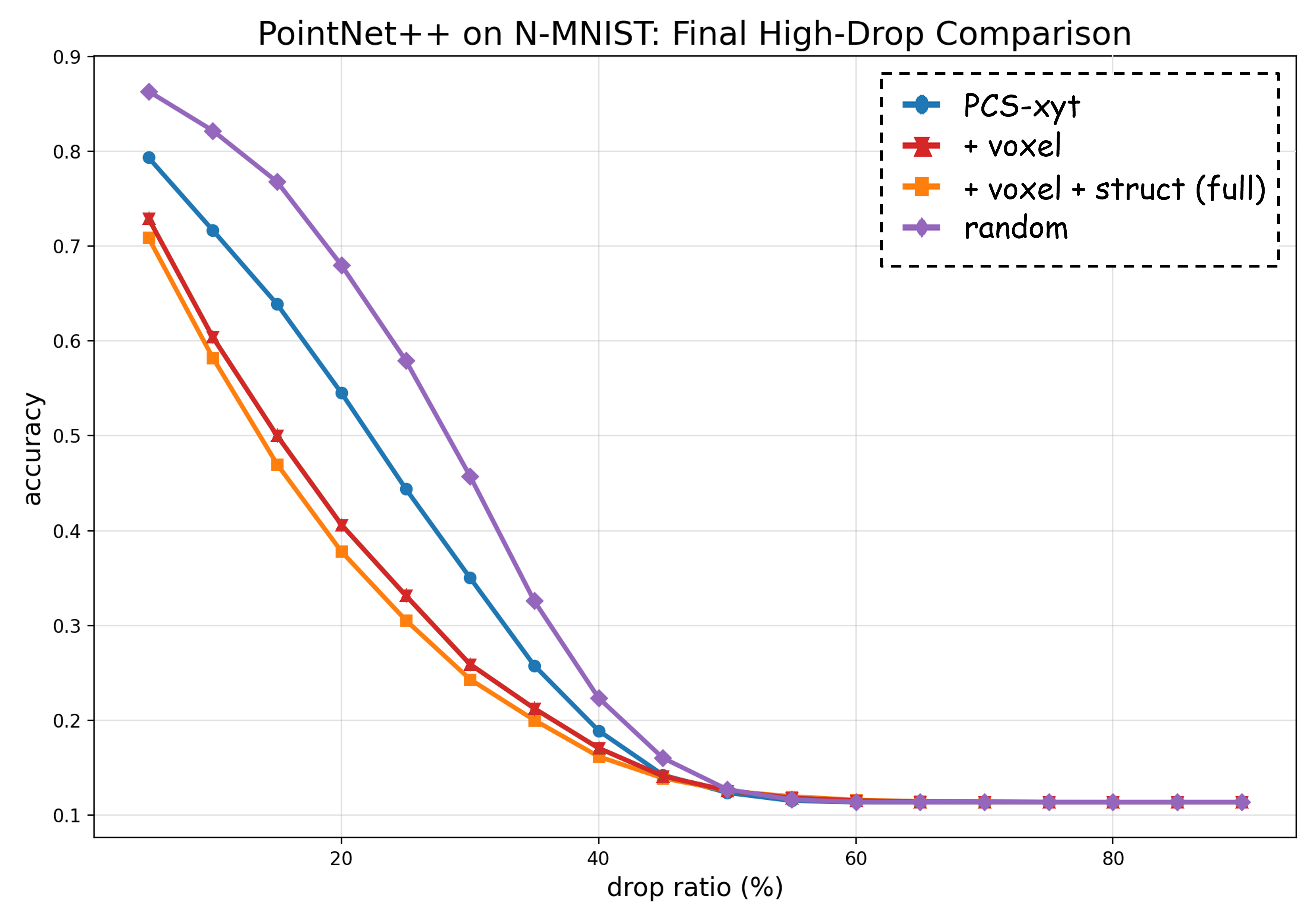}

\vspace{-0.3em}

\includegraphics[width=0.82\columnwidth]{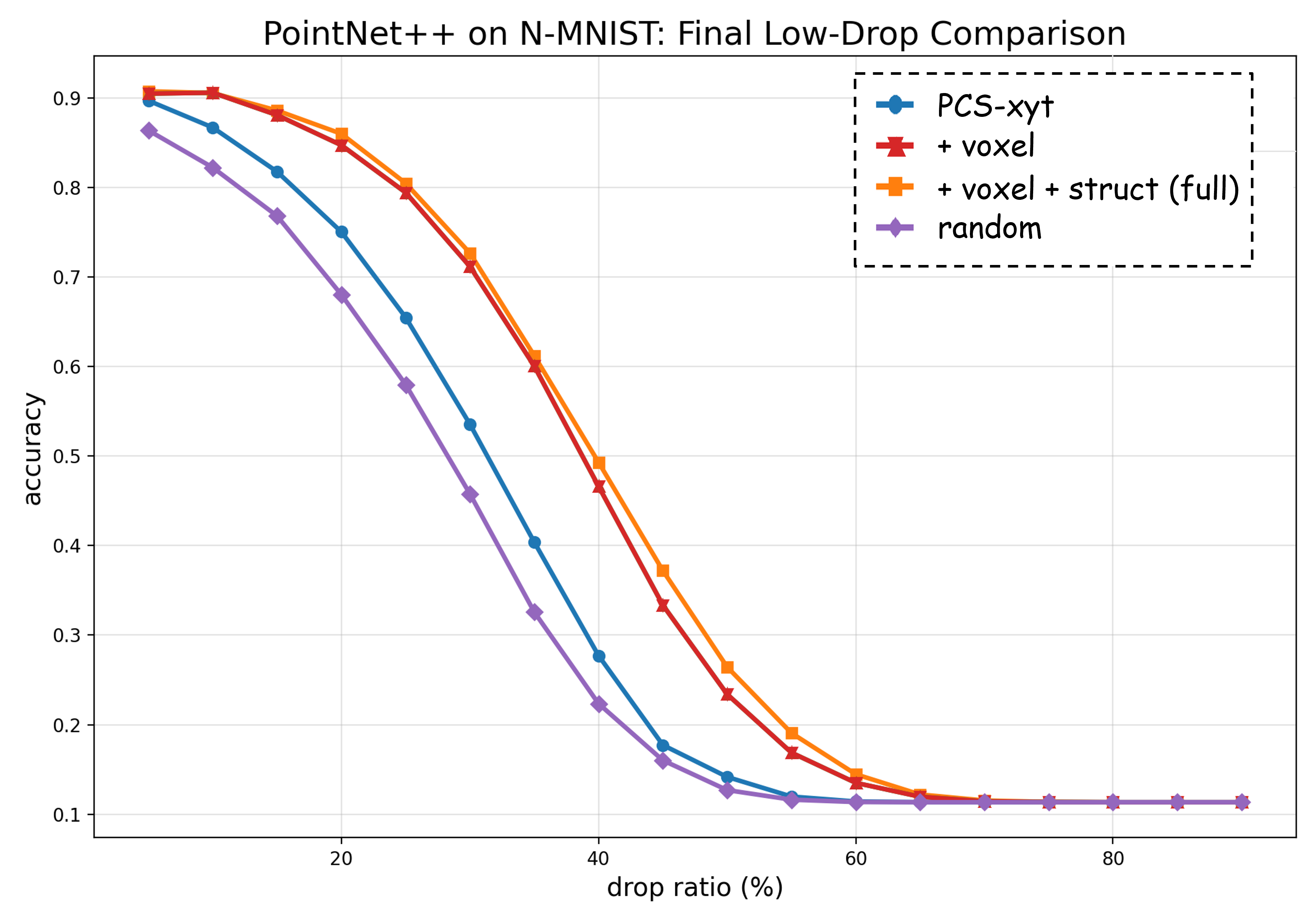}

\vspace{-0.8em}
\caption{Component ablation on the PointNet++/N-MNIST development subset. Top: high-tail deletion AUC (lower is better). Bottom: low-tail deletion AUC (higher is better).}
\label{fig:ablation}
\vspace{-1.2em}
\end{figure}

We ablate the key evidence channels on the deterministic PointNet++/N-MNIST development subset using the same 5\%--90\% deletion protocol. Figure~\ref{fig:ablation} compares four representative rankings: \pcs, \pcs with voxel evidence, the full voxel-structural fusion, and Random. The top panel reports high-tail deletion, where lower accuracy indicates that the removed events are more decision-critical. The bottom panel reports low-tail deletion, where higher accuracy indicates that the removed events are less important.

The high-tail results show that voxel evidence is the main source of decision-critical localization. Compared with \pcs, adding the voxel channel causes accuracy to drop earlier across most deletion ratios. The full fusion, which further incorporates structural ranking, remains among the strongest high-tail curves and stays well below Random, indicating that it preserves the ability to identify influential events. The low-tail results reveal the complementary role of the structural component. While voxel evidence alone improves high-tail deletion, the full voxel-structural fusion gives the best low-tail preservation over most ratios, keeping accuracy higher than \pcs, voxel-only, and Random before the curves converge at large deletion ratios. Together, the two panels show that voxel perturbation sharpens the head of the ranking, while structural fusion improves the tail. This supports the design of \method: combining regional task evidence with global structural ordering yields a ranking that is effective for both explanation and safe event removal.





\subsection{Representative Prediction Changes}

\begin{figure}[t]
\centering
\includegraphics[width=0.98\columnwidth]{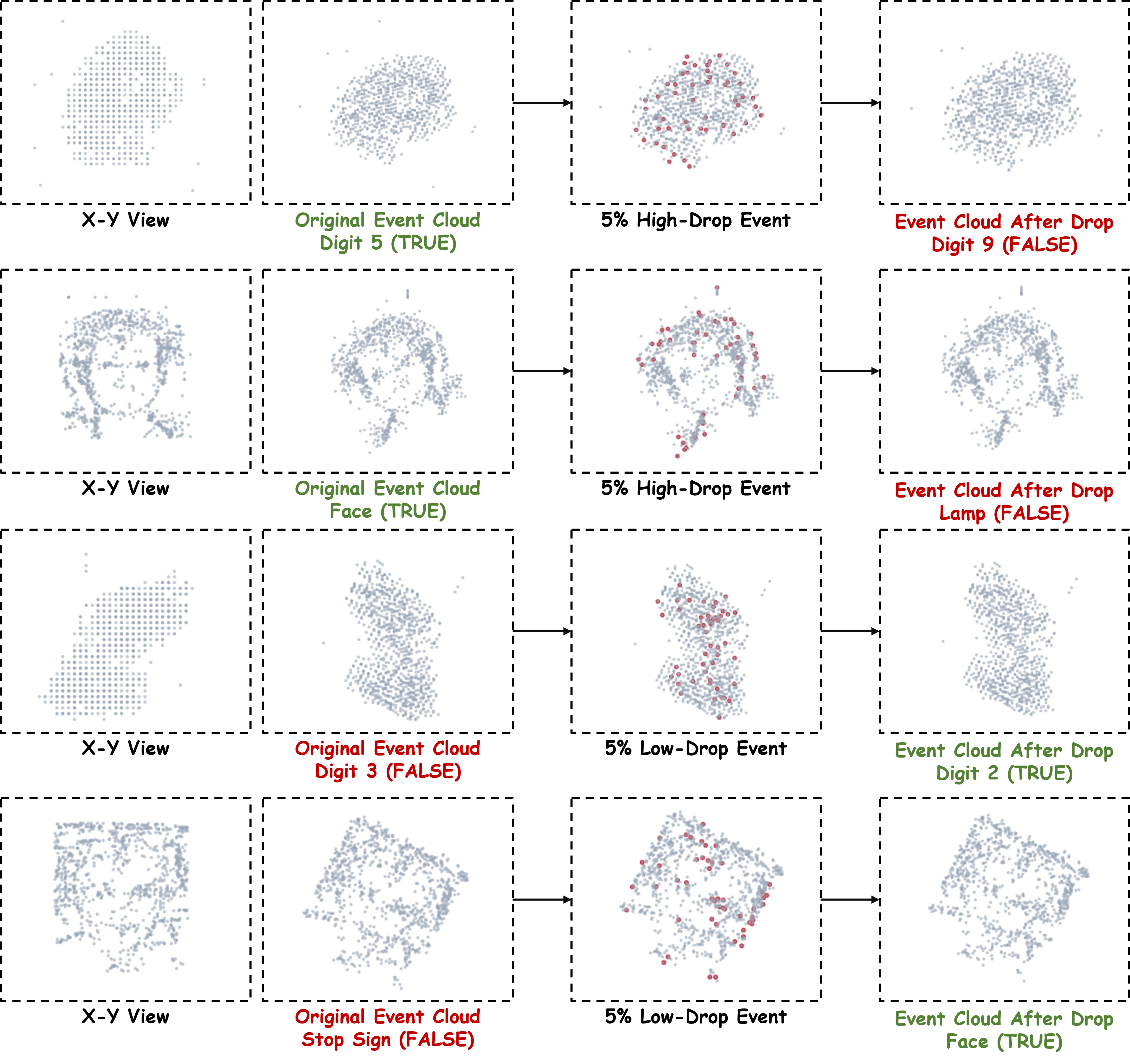}
\caption{Representative prediction changes after deleting ranked events under \method. Red points denote deleted events; green/red text marks correct/incorrect predictions.}
\label{fig:qualitative}
\end{figure}

Figure~\ref{fig:qualitative} shows representative prediction changes caused by deleting events selected by \method. Each row contains the original event cloud, the deleted events highlighted in red, and the retained event cloud after deletion. Green labels indicate correct predictions, while red labels indicate incorrect predictions.

The first two rows show failure-inducing high-tail deletions. Removing a small set of highly ranked events changes originally correct predictions into incorrect ones, from digit 5 to digit 9 and from face to lamp. These cases suggest that the removed events contain decision-critical evidence for the original class.

The last two rows show the complementary behavior. For samples that are initially misclassified, deleting ranked events suppresses misleading evidence and recovers the correct class, changing digit 3 to digit 2 and stop sign to face. Overall, these examples connect the aggregate two-tail deletion curves with concrete event-level behavior: the ranking identifies influential events whose removal can alter or even correct the model decision.

\section{Conclusion}


We presented \method, a training-free method that fuses event-level gradient evidence with task-aware voxel perturbations to produce a global event ranking for two-tail deletion. Across nine archived dataset--backbone settings, \method improves both tails over \pcs in every case. Its head identifies decision-critical events, while its tail identifies removable events that preserve accuracy, so one ranking supports both explanation and compression. The two-tail deletion protocol is architecture-agnostic and reproducible from the archived evidence trail. 
Beyond individual model interpretation, event cloud attribution provides a general tool for analyzing how asynchronous event streams contribute to perception decisions. Such event-level evidence may further support future research in architecture design, adversarial robustness, event selection, and efficient event processing.

{\small
\setlength{\bibsep}{2.0pt plus 0.5pt}
\bibliography{references}

@article{gallego2022event,
  title={Event-based vision: A survey},
  author={Gallego, Guillermo and Delbr{\"u}ck, Tobias and Orchard, Garrick Michael and Bartolozzi, Chiara and Taba, Brian and Censi, Andrea and Leutenegger, Stefan and Davison, Andrew J and Conradt, J{\"o}rg and Daniilidis, Kostas and others},
  journal={IEEE transactions on pattern analysis and machine intelligence},
  volume={44},
  number={1},
  pages={154--180},
  year={2022},
  publisher={IEEE}
}

@inproceedings{gehrig2019end,
  title={End-to-end learning of representations for asynchronous event-based data},
  author={Gehrig, Daniel and Loquercio, Antonio and Derpanis, Konstantinos G and Scaramuzza, Davide},
  booktitle={Proceedings of the IEEE/CVF international conference on computer vision},
  pages={5633--5643},
  year={2019}
}

@inproceedings{sironi2018hats,
  title={HATS: Histograms of averaged time surfaces for robust event-based object classification},
  author={Sironi, Amos and Brambilla, Manuele and Bourdis, Nicolas and Lagorce, Xavier and Benosman, Ryad},
  booktitle={Proceedings of the IEEE conference on computer vision and pattern recognition},
  pages={1731--1740},
  year={2018}
}

@inproceedings{sabater2022event,
  title={Event transformer. a sparse-aware solution for efficient event data processing},
  author={Sabater, Alberto and Montesano, Luis and Murillo, Ana C},
  booktitle={Proceedings of the IEEE/CVF Conference on Computer Vision and Pattern Recognition},
  pages={2677--2686},
  year={2022}
}

@inproceedings{peng2023get,
  title={Get: Group event transformer for event-based vision},
  author={Peng, Yansong and Zhang, Yueyi and Xiong, Zhiwei and Sun, Xiaoyan and Wu, Feng},
  booktitle={Proceedings of the IEEE/CVF International Conference on Computer Vision},
  pages={6038--6048},
  year={2023}
}

@inproceedings{qi2017pointnet,
  title={Pointnet: Deep learning on point sets for 3d classification and segmentation},
  author={Qi, Charles R and Su, Hao and Mo, Kaichun and Guibas, Leonidas J},
  booktitle={Proceedings of the IEEE conference on computer vision and pattern recognition},
  pages={652--660},
  year={2017}
}

@article{qi2017pointnet++,
  title={Pointnet++: Deep hierarchical feature learning on point sets in a metric space},
  author={Qi, Charles Ruizhongtai and Yi, Li and Su, Hao and Guibas, Leonidas J},
  journal={Advances in neural information processing systems},
  volume={30},
  year={2017}
}

@article{ren2025rethinking,
  title={Rethinking efficient and effective point-based networks for event camera classification and regression},
  author={Ren, Hongwei and Zhou, Yue and Zhu, Jiadong and Lin, Xiaopeng and Fu, Haotian and Huang, Yulong and Fang, Yuetong and Ma, Fei and Yu, Hao and Cheng, Bojun},
  journal={IEEE Transactions on Pattern Analysis and Machine Intelligence},
  year={2025},
  publisher={IEEE}
}

@article{lin2023e2pnet,
  title={E2pnet: event to point cloud registration with spatio-temporal representation learning},
  author={Lin, Xiuhong and Qiu, Changjie and Shen, Siqi and Zang, Yu and Liu, Weiquan and Bian, Xuesheng and M{\"u}ller, Matthias and Wang, Cheng and others},
  journal={Advances in Neural Information Processing Systems},
  volume={36},
  pages={18076--18089},
  year={2023}
}

@inproceedings{ren2024simple,
  title={A simple and effective point-based network for event camera 6-dofs pose relocalization},
  author={Ren, Hongwei and Zhu, Jiadong and Zhou, Yue and Fu, Haotian and Huang, Yulong and Cheng, Bojun},
  booktitle={Proceedings of the IEEE/CVF Conference on Computer Vision and Pattern Recognition},
  pages={18112--18121},
  year={2024}
}

@inproceedings{zheng2019pointcloud,
  title={Pointcloud saliency maps},
  author={Zheng, Tianhang and Chen, Changyou and Yuan, Junsong and Li, Bo and Ren, Kui},
  booktitle={Proceedings of the IEEE/CVF international conference on computer vision},
  pages={1598--1606},
  year={2019}
}

@article{simonyan2013deep,
  title={Deep inside convolutional networks: Visualising image classification models and saliency maps},
  author={Simonyan, Karen and Vedaldi, Andrea and Zisserman, Andrew},
  journal={arXiv preprint arXiv:1312.6034},
  year={2013}
}

@inproceedings{sundararajan2017axiomatic,
  title={Axiomatic attribution for deep networks},
  author={Sundararajan, Mukund and Taly, Ankur and Yan, Qiqi},
  booktitle={International conference on machine learning},
  pages={3319--3328},
  year={2017},
  organization={PMLR}
}

@inproceedings{selvaraju2017grad,
  title={Grad-cam: Visual explanations from deep networks via gradient-based localization},
  author={Selvaraju, Ramprasaath R and Cogswell, Michael and Das, Abhishek and Vedantam, Ramakrishna and Parikh, Devi and Batra, Dhruv},
  booktitle={Proceedings of the IEEE international conference on computer vision},
  pages={618--626},
  year={2017}
}

@article{petsiuk2018rise,
  title={Rise: Randomized input sampling for explanation of black-box models},
  author={Petsiuk, Vitali and Das, Abir and Saenko, Kate},
  journal={arXiv preprint arXiv:1806.07421},
  year={2018}
}

@inproceedings{fong2017interpretable,
  title={Interpretable explanations of black boxes by meaningful perturbation},
  author={Fong, Ruth C and Vedaldi, Andrea},
  booktitle={Proceedings of the IEEE international conference on computer vision},
  pages={3429--3437},
  year={2017}
}

@article{adebayo2018sanity,
  title={Sanity checks for saliency maps},
  author={Adebayo, Julius and Gilmer, Justin and Muelly, Michael and Goodfellow, Ian and Hardt, Moritz and Kim, Been},
  journal={Advances in neural information processing systems},
  volume={31},
  year={2018}
}

@inproceedings{tomsett2020sanity,
  title={Sanity checks for saliency metrics},
  author={Tomsett, Richard and Harborne, Dan and Chakraborty, Supriyo and Gurram, Prudhvi and Preece, Alun},
  booktitle={Proceedings of the AAAI conference on artificial intelligence},
  volume={34},
  number={04},
  pages={6021--6029},
  year={2020}
}

@article{yeh2019fidelity,
  title={On the (in) fidelity and sensitivity of explanations},
  author={Yeh, Chih-Kuan and Hsieh, Cheng-Yu and Suggala, Arun and Inouye, David I and Ravikumar, Pradeep K},
  journal={Advances in neural information processing systems},
  volume={32},
  year={2019}
}

@article{hooker2019benchmark,
  title={A benchmark for interpretability methods in deep neural networks},
  author={Hooker, Sara and Erhan, Dumitru and Kindermans, Pieter-Jan and Kim, Been},
  journal={Advances in neural information processing systems},
  volume={32},
  year={2019}
}

@inproceedings{gomez2022metrics,
  title={Metrics for saliency map evaluation of deep learning explanation methods},
  author={Gomez, Tristan and Fr{\'e}our, Thomas and Mouch{\`e}re, Harold},
  booktitle={International Conference on Pattern Recognition and Artificial Intelligence},
  pages={84--95},
  year={2022},
  organization={Springer}
}

@article{levi2024fast,
  title={Fast and simple explainability for point cloud networks},
  author={Levi, Meir Yossef and Gilboa, Guy},
  journal={arXiv preprint arXiv:2403.07706},
  year={2024}
}

@article{ahmadi2024explainability,
  title={Explainability of point cloud neural networks using smile: Statistical model-agnostic interpretability with local explanations},
  author={Ahmadi, Seyed Mohammad and Aslansefat, Koorosh and Valcarce-Di{\~n}eiro, Rub{\'e}n and Barnfather, Joshua},
  journal={arXiv preprint arXiv:2410.15374},
  year={2024}
}

@article{mulawade2025xai,
  title={XAI for Point Cloud Data Using Perturbations Based on Meaningful Segmentation},
  author={Mulawade, Raju Ningappa and Garth, Christoph and Wiebel, Alexander},
  journal={IEEE Access},
  year={2025},
  publisher={IEEE}
}

@inproceedings{huang2023eventpoint,
  title={Eventpoint: Self-supervised interest point detection and description for event-based camera},
  author={Huang, Ze and Sun, Li and Zhao, Cheng and Li, Song and Su, Songzhi},
  booktitle={Proceedings of the IEEE/CVF Winter Conference on Applications of Computer Vision},
  pages={5396--5405},
  year={2023}
}

@inproceedings{feng2024interpretable3d,
  title={Interpretable3d: An ad-hoc interpretable classifier for 3d point clouds},
  author={Feng, Tuo and Quan, Ruijie and Wang, Xiaohan and Wang, Wenguan and Yang, Yi},
  booktitle={Proceedings of the AAAI Conference on Artificial Intelligence},
  volume={38},
  number={2},
  pages={1761--1769},
  year={2024}
}

@inproceedings{amir2017low,
  title={A low power, fully event-based gesture recognition system},
  author={Amir, Arnon and Taba, Brian and Berg, David and Melano, Timothy and McKinstry, Jeffrey and Di Nolfo, Carmelo and Nayak, Tapan and Andreopoulos, Alexander and Garreau, Guillaume and Mendoza, Marcela and others},
  booktitle={Proceedings of the IEEE conference on computer vision and pattern recognition},
  pages={7243--7252},
  year={2017}
}

@article{orchard2015converting,
  title={Converting static image datasets to spiking neuromorphic datasets using saccades},
  author={Orchard, Garrick and Jayawant, Ajinkya and Cohen, Gregory K and Thakor, Nitish},
  journal={Frontiers in neuroscience},
  volume={9},
  pages={437},
  year={2015},
  publisher={Frontiers Media SA}
}

@article{bitar2023gradient,
  title={Gradient-based feature-attribution explainability methods for spiking neural networks},
  author={Bitar, Ammar and Rosales, Rafael and Paulitsch, Michael},
  journal={Frontiers in Neuroscience},
  volume={17},
  pages={1153999},
  year={2023},
  publisher={Frontiers Media SA}
}

@article{liu2024ffam,
  title={FFAM: feature factorization activation map for explanation of 3d detectors},
  author={Liu, Shuai and Li, Boyang and Fang, Zhiyu and Cui, Mingyue and Huang, Kai},
  journal={Advances in Neural Information Processing Systems},
  volume={37},
  pages={80363--80382},
  year={2024}
}

@article{ren2026systematic,
  title={A Systematic Survey on Event Camera Representation Learning},
  author={Ren, Hongwei and Jiang, Youxin and Huang, Tuopusen and Wu, Xiangqian},
  journal={arXiv preprint arXiv:2606.23078},
  year={2026}
}

@inproceedings{cannici2020differentiable,
  title={A differentiable recurrent surface for asynchronous event-based data},
  author={Cannici, Marco and Ciccone, Marco and Romanoni, Andrea and Matteucci, Matteo},
  booktitle={European Conference on Computer Vision},
  pages={136--152},
  year={2020},
  organization={Springer}
}

@inproceedings{sekikawa2019eventnet,
  title={Eventnet: Asynchronous recursive event processing},
  author={Sekikawa, Yusuke and Hara, Kosuke and Saito, Hideo},
  booktitle={Proceedings of the IEEE/CVF conference on computer vision and pattern recognition},
  pages={3887--3896},
  year={2019}
}

@inproceedings{schaefer2022aegnn,
  title={Aegnn: Asynchronous event-based graph neural networks},
  author={Schaefer, Simon and Gehrig, Daniel and Scaramuzza, Davide},
  booktitle={Proceedings of the IEEE/CVF conference on computer vision and pattern recognition},
  pages={12371--12381},
  year={2022}
}

@inproceedings{deng2022voxel,
  title={A voxel graph cnn for object classification with event cameras},
  author={Deng, Yongjian and Chen, Hao and Liu, Hai and Li, Youfu},
  booktitle={Proceedings of the IEEE/CVF Conference on Computer Vision and Pattern Recognition},
  pages={1172--1181},
  year={2022}
}

@inproceedings{ren2026scalable,
  title={Scalable Event Cloud Network for Event-based Classification},
  author={Ren, Hongwei and Ma, Fei and Fang, Yuetong and Huang, Hongxiang and Zhou, Yue and Huang, Yulong and FU, Haotian and Yang, Ziyi and Jiang, Youxin and Wu, Xiangqian and others},
  booktitle={Forty-third International Conference on Machine Learning},
  year={2026}
}

@inproceedings{ribeiro2016should,
  title={" Why should i trust you?" Explaining the predictions of any classifier},
  author={Ribeiro, Marco Tulio and Singh, Sameer and Guestrin, Carlos},
  booktitle={Proceedings of the 22nd ACM SIGKDD international conference on knowledge discovery and data mining},
  pages={1135--1144},
  year={2016}
}

@article{lundberg2017unified,
  title={A unified approach to interpreting model predictions},
  author={Lundberg, Scott M and Lee, Su-In},
  journal={Advances in neural information processing systems},
  volume={30},
  year={2017}
}

@article{bach2015pixel,
  title={On pixel-wise explanations for non-linear classifier decisions by layer-wise relevance propagation},
  author={Bach, Sebastian and Binder, Alexander and Montavon, Gr{\'e}goire and Klauschen, Frederick and M{\"u}ller, Klaus-Robert and Samek, Wojciech},
  journal={PloS one},
  volume={10},
  number={7},
  pages={e0130140},
  year={2015},
  publisher={Public Library of Science San Francisco, CA USA}
}

@article{smilkov2017smoothgrad,
  title={Smoothgrad: removing noise by adding noise},
  author={Smilkov, Daniel and Thorat, Nikhil and Kim, Been and Vi{\'e}gas, Fernanda and Wattenberg, Martin},
  journal={arXiv preprint arXiv:1706.03825},
  year={2017}
}

@inproceedings{chattopadhay2018grad,
  title={Grad-cam++: Generalized gradient-based visual explanations for deep convolutional networks},
  author={Chattopadhay, Aditya and Sarkar, Anirban and Howlader, Prantik and Balasubramanian, Vineeth N},
  booktitle={2018 IEEE winter conference on applications of computer vision (WACV)},
  pages={839--847},
  year={2018},
  organization={IEEE}
}

@inproceedings{zeiler2014visualizing,
  title={Visualizing and understanding convolutional networks},
  author={Zeiler, Matthew D and Fergus, Rob},
  booktitle={European conference on computer vision},
  pages={818--833},
  year={2014},
  organization={Springer}
}

@incollection{kindermans2019reliability,
  title={The (un) reliability of saliency methods},
  author={Kindermans, Pieter-Jan and Hooker, Sara and Adebayo, Julius and Alber, Maximilian and Sch{\"u}tt, Kristof T and D{\"a}hne, Sven and Erhan, Dumitru and Kim, Been},
  booktitle={Explainable AI: Interpreting, explaining and visualizing deep learning},
  pages={267--280},
  year={2019},
  publisher={Springer}
}

@article{alvarez2018robustness,
  title={On the robustness of interpretability methods},
  author={Alvarez-Melis, David and Jaakkola, Tommi S},
  journal={arXiv preprint arXiv:1806.08049},
  year={2018}
}

@article{wang2019dynamic,
  title={Dynamic graph cnn for learning on point clouds},
  author={Wang, Yue and Sun, Yongbin and Liu, Ziwei and Sarma, Sanjay E and Bronstein, Michael M and Solomon, Justin M},
  journal={ACM Transactions on Graphics (tog)},
  volume={38},
  number={5},
  pages={1--12},
  year={2019},
  publisher={Acm New York, NY, USA}
}

@inproceedings{zhao2021point,
  title={Point transformer},
  author={Zhao, Hengshuang and Jiang, Li and Jia, Jiaya and Torr, Philip HS and Koltun, Vladlen},
  booktitle={Proceedings of the IEEE/CVF international conference on computer vision},
  pages={16259--16268},
  year={2021}
}

@article{li2017cifar10,
  title={Cifar10-dvs: an event-stream dataset for object classification},
  author={Li, Hongmin and Liu, Hanchao and Ji, Xiangyang and Li, Guoqi and Shi, Luping},
  journal={Frontiers in neuroscience},
  volume={11},
  pages={244131},
  year={2017},
  publisher={Frontiers}
}

@inproceedings{ren2024spikepoint,
  title={Spikepoint: An efficient point-based spiking neural network for event cameras action recognition},
  author={Ren, Hongwei and Zhou, Yue and Lin, Xiaopeng and Huang, Yulong and Fu, Haotian and Song, Jie and Cheng, Bojun},
  booktitle={International Conference on Learning Representations},
  volume={2024},
  pages={27827--27846},
  year={2024}
}
}

\end{document}